\documentclass[pmlr]{jmlr}

\usepackage{longtable}
\usepackage{subcaption} 
\usepackage{float}
\usepackage{hyperref}
\usepackage{booktabs}
\usepackage[load-configurations=version-1]{siunitx} 
\usepackage{comment}

\theorembodyfont{\upshape}
\theoremheaderfont{\scshape}
\theorempostheader{:}
\theoremsep{\newline}

\jmlrvolume{266}
\jmlryear{2025}
\jmlrworkshop{Conformal and Probabilistic Prediction with Applications}
\jmlrproceedings{PMLR}{Proceedings of Machine Learning Research}

\title{Robust Vision-Based Runway Detection through Conformal Prediction and Conformal mAP}

\author{
  \Name{Alya Zouzou}\Email{alyasltd@gmail.com}\\
  \addr Paul Valery University, Montpellier III, France\\
  Airbus, Toulouse, France
  \AND
  \Name{Léo Andéol}\Email{leo.andeol@math.univ-toulouse.fr}\\
  \addr Institut de Mathématiques de Toulouse, Toulouse, France\\
  SNCF, Saint-Denis, France
  \AND
  \Name{Mélanie Ducoffe}\Email{melanie.ducoffe@airbus.com}\\
  \addr Airbus, Toulouse, France\\
  IRT Saint Exupéry, Toulouse, France
  \AND
  \Name{Ryma Boumazouza}\Email{ryma.boumazouza@airbus.com}\\
  \addr Airbus, Toulouse, France\\
  IRT Saint Exupéry, Toulouse, France
}

\begin{document}

\maketitle

\begin{abstract}
We explore the use of conformal prediction to provide statistical uncertainty guarantees for runway detection in vision-based landing systems (VLS). Using fine-tuned YOLOv5 and YOLOv6 models on aerial imagery, we apply conformal prediction to quantify localization reliability under user-defined risk levels. We also introduce \textbf{Conformal mean Average Precision} (C-mAP), a novel metric aligning object detection performance with conformal guarantees.
Our results show that conformal prediction can improve the reliability of runway detection by quantifying uncertainty in a statistically sound way, increasing safety on-board and paving the way for certification of ML system in the aerospace domain.
\end{abstract}

\begin{figure}[htbp]
\floatconts
  {fig:pipeline}
  {\caption{Illustration of conformal prediction on the runway detection task. \textcolor{red}{Red: Ground Truth}, \textcolor{blue}{Blue: YOLO Prediction}, \textcolor{green}{Green: Outer Conformalized Box}.
}}
  {\includegraphics[width=0.5\linewidth]{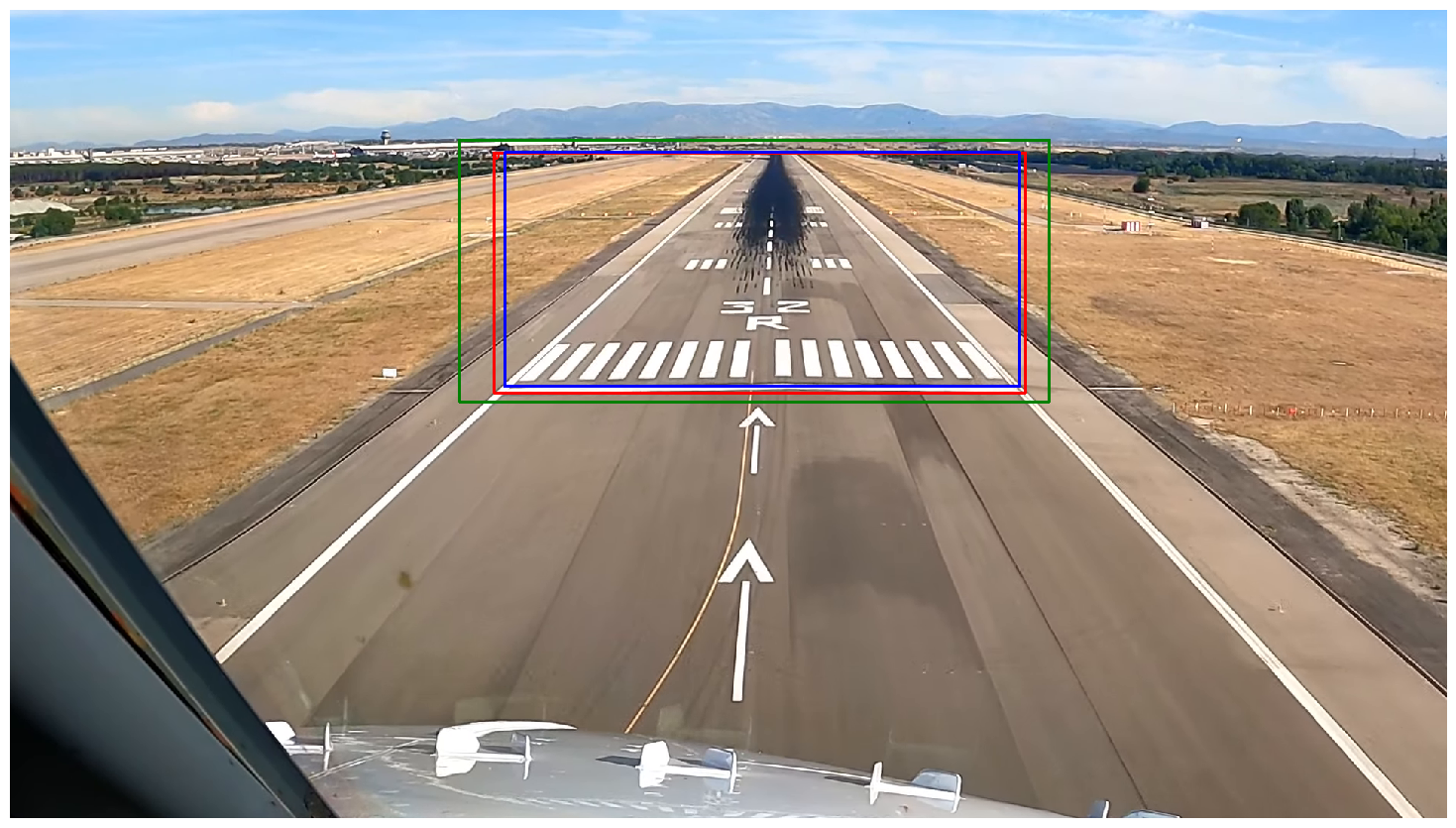}}
\end{figure}

\section{Introduction}
\label{sec:introduction}

The emergence of Machine Learning (ML), particularly deep learning and neural network (NN) models, has enabled transformative progress across a wide range of domains including transportation, healthcare, and finance. 
However, despite their impressive performance, neural networks exhibit inherent vulnerabilities, most notably their sensitivity to adversarial inputs and data distribution shifts, as highlighted in foundational work such as \cite{szegedy2013intriguing}. These weaknesses raise critical concerns when deploying ML systems in safety-critical contexts, where guarantees on reliability, robustness, and uncertainty quantification are paramount.

One of the most stringent and high-stakes domains for ML deployment is aerospace. Within this field, vision-based landing systems (VLS) have emerged as a promising application of ML to enhance or potentially replace traditional landing aids. Aircraft landing is among the most complex and risk-prone phases of flight, currently supported by redundant human and technical systems—two-pilot operations, standardized communication protocols (e.g., ATIS), and ground-based navigation aids such as the Instrument Landing System (ILS) and GPS. As the industry moves toward single-pilot operations and increasing automation, ML-based systems must be able to provide reliable, interpretable, and certifiable support for tasks like visual runway detection and localization.

We explore recent developments in conformal prediction as a tool for uncertainty quantification (UQ) in object detection models applied to aerial imagery. These tools offer statistical guarantees on localization coverage and can help satisfy regulatory demands regarding AI in aerospace.
Specifically, we focus on the core task of runway detection, which serves as a critical anchor for downstream aircraft pose estimation. Accurate detection of the runway is foundational—errors at this stage propagate through the pipeline and could compromise the entire landing guidance system.

Our contributions are as follows:

\begin{itemize}
    \item We apply and evaluate conformal prediction with different margins over boxes (e.g., additive or multiplicative penalties) on multiple object detection architectures for the task of runway detection in aerial images.
    \item We introduce a novel evaluation metric, the conformal mean Average Precision (C-mAP), which quantifies the trade-off between accurate bounding box predictions and their coverage of ground-truth boxes at user-defined risk levels—thus aligning conformal prediction with established object detection benchmarks.
    \item We release our codebase \footnote{\url{https://github.com/alyasltd/conformal_runway_detection}}, trained models, and experimental protocols as an open-source repository to encourage reproducibility and further research in this critical area.
\end{itemize}

\section{Object Detection Overview}

Object detection refers to the task of identifying and localizing objects within an image, along with classifying them into predefined categories. Recent works in object detection show a great variety of models, mostly performed by neural networks, which can be categorized into two main types: \textbf{two-stage detectors} and \textbf{single-stage detectors}. 

\begin{itemize}
    \item \textbf{Two-stage detectors} first identify Regions of Interest (RoIs) in the image, and then classify those regions as containing or not objects (R-CNN, Mask R-CNN, \cite{bharati2020deep}).
    \item \textbf{Single-stage detectors}, such as the YOLO (You Only Look Once, \cite{terven2023comprehensive}) family, are known for their efficiency and trade-off between speed and accuracy. They directly predict bounding boxes and class probabilities for each object in a single pass through the network. Other examples of single-stage detectors include DETR and FCOS \cite{carion2020end, tian2020fcos}.
\end{itemize}

The goal of object detection can be formally defined as follows: given an image containing potential objects from a target set of classes, the object detector must localize each object and predict its class label. Localization is achieved through the use of bounding boxes. Each predicted bounding box is associated with the following information:

\begin{itemize}
    \item \textbf{Localization vector}: a vector denoted as $\bold{\hat{b}}=[x_{min}, y_{min}, x_{max}, y_{max}]$, which are typically represented by their top-left and bottom-right coordinates.
    \item \textbf{Objectness score} $o$: A probability value between 0 and 1, representing the model's confidence that an object exists within the bounding box, regardless of its class label. 
    \item \textbf{Class probability vector} $\mathbf{p}$: A vector containing class probabilities, where the highest value indicates the predicted class of the object, provided that the objectness score exceeds a certain threshold (indicating that the object exists).
\end{itemize}

From these two last sources of information, we compute the \textbf{confidence score}, defined as the product of the objectness score and the class probability. Although this score is sometimes used as a proxy for uncertainty estimation \cite{wenkel2021confidence}, it lacks reliability—often over- or underestimating the true uncertainty—and provides no statistical guarantees.

In practice, object detectors can produce thousands of bounding boxes for a single image, many of which may significantly overlap. This overlap is typically quantified using the \textit{Intersection over Union (IoU)} metric, defined as the area of the intersection divided by the area of the union of the predicted and ground truth bounding boxes:
\[
\text{IoU} = \frac{\text{Area of Intersection}}{\text{Area of Union}}
\]

To handle such redundancy, a post-processing step is applied, most commonly the Non-Maximum Suppression (NMS, \cite{neubeck2006efficient}) algorithm, which removes overlapping bounding boxes.
A second step consists in filtering out insufficiently confident boxes of the final sequence of prediction boxes.

We define an object detection model $f$ as the composition of the object detector (after NMS) and confidence threshold. Formally, we start with a  model $f^\mathrm{NMS}$ taking as input an image $x$ and returning a sequence of fixed length $N$. These predictions are then filtered for (when two boxes shared an IoU over $t$, the least confident is disregarded). This results in a sequence of predictions of variable length $N^\mathrm{NMS}_x$ formally denoted by 

\[
f^\mathrm{NMS}: x \rightarrow \left(\mathbf{\hat{b}}_i, o_i, \mathbf{p}_i \right)_{i=1}^{N^\mathrm{NMS}_x},
\]

where $\mathbf{\hat{b}}_i$ is the bounding box for the $i$-th prediction, $o_i$ is the corresponding objectness score, $\mathbf{p}_i$ is the vector of class probabilities.

Finally, all predictions with objectness scores above a threshold $\tau$ are preserved,

\[
    I(x) = \left\{i \in \{1,\dots,N^\mathrm{NMS}_x\} : \ o_{[i]} \geq \tau\right\},
\]

where $o_{[i]}$ is the $i$-th largest objectness score in $f(x)$. We then obtain the final predictions

\[
    f : x \rightarrow \left\{f^\mathrm{NMS}(x)_{[k]}, \ k\in I(x)\right\}.
\]

The threshold $\tau$ is a predefined value from the filtering stage used to discard predictions with low objectness scores, ensuring that only detections with sufficiently high objectness score are considered.

\subsection{Model Evaluation: Mean Average Precision (mAP)}

The performance of object detection models is commonly evaluated using the \textbf{mean Average Precision (mAP)} metric, which is the mean of the \textbf{Average Precision (AP)} values computed across all object classes.

Consider the set of predicted bounding boxes $\{\mathbf{\hat{b}}^x_i\}$ output by the object detector, and the ground truth annotations be $\{(\mathbf{b}^{x}_j, c^x_j)\}_j$, where $\mathbf{b}^{x}_j$ is a ground truth box and $c^x_j$ its corresponding class label. Denoting $C$ as the total number of classes, mAP is defined as:

\[
\text{mAP} = \frac{1}{C} \sum_c AP\left(\{\mathbf{\hat{b}}^x_i\}_{\hat{c}_i^x = c}, \{\mathbf{b}^{x}_j\}_{c^x_j = c}\right),
\]
where $\hat{c}_i^x=\arg\max_k \mathbf{p}_i^x(k)$.

We now focus on the definition of \textbf{Average Precision (AP)}. To compute AP, predicted bounding boxes are matched per image to ground truth boxes using a similarity measure, typically the \textit{Intersection over Union (IoU)}. A predicted box is considered a \textbf{true positive} if its IoU with a ground truth box exceeds a predefined threshold and the class label matches; otherwise, it is a \textbf{false positive}. Ground truth boxes not matched to any prediction are considered \textbf{false negatives}.

Predicted boxes are ranked from highest to lowest confidence score (excluding those below a minimum confidence threshold), and this ranked list is used to construct a \textbf{precision-recall curve}. Recall that:
\begin{itemize}
    \item \textbf{Precision} is the cumulative ratio of true positives to the sum of true positives and false positives:
    \[
    \text{Precision}: p = \frac{\text{TP}}{\text{TP} + \text{FP}}
    \]
    \item \textbf{Recall} is the cumulative ratio of true positives to the sum of true positives and false negatives:
    \[
    \text{Recall}: r= \frac{\text{TP}}{\text{TP} + \text{FN}}
    \]
\end{itemize}

The precision-recall curve plots \textit{precision as a function of
recall}, denoted as \( p(r) \), where \( p(r) \) represents the precision 
achieved at a given level of recall \( r \). The \textbf{average precision} 
for a class is then computed as the area under this precision-recall curve. This curve does not follow a specific shape, which can make the computation 
of its area expensive. To simplify this, a preprocessing step is applied to 
overestimate the area by assigning to each recall value the highest 
precision observed at any greater recall value. We denote this interpolated 
precision as:

\[
\overline{p}(r) = \max_{\tilde{r} \geq r} p(\tilde{r})
\]

To ensure a more stable and representative measurement, the precision-recall curve is typically interpolated. One common method is the \textit{11-point interpolation}, where the precision at recall levels $\{0.0, 0.1, \ldots, 1.0\}$. More modern implementation such as those used in the COCO evaluation protocol may involve more interpolation points. Finally the AP is then computed as the average of these interpolated precision values:
\[
AP = \frac{1}{11} \sum_{r \in \{0.0, 0.1, \ldots, 1.0\}} \overline{p}(r)
\]

The mAP computed using an IoU threshold $t$ is denoted as \textbf{mAP@t}. The most commonly reported metrics are: (i) \textbf{mAP@50}: mAP with IoU threshold set to 0.5
(ii) \textbf{mAP@50:95}: mAP averaged over IoU thresholds ranging from 0.5 to 0.95, in steps of 0.05

\begin{figure}[h]
  \centering
  \includegraphics[width=\linewidth]{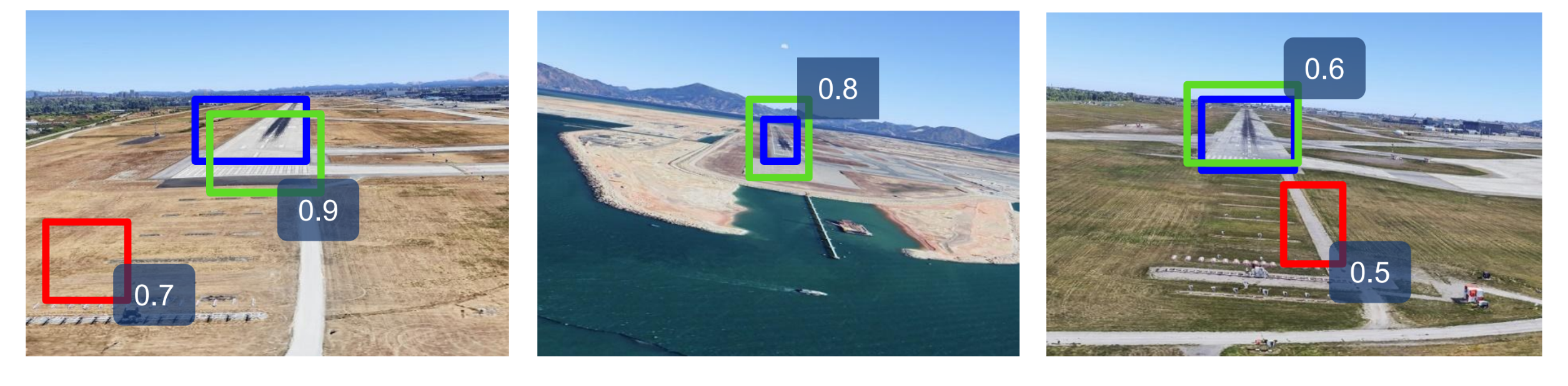}

  \vspace{0.5em} 

  \begin{tabular}{>{\centering\arraybackslash}m{3.5cm} 
                  >{\centering\arraybackslash}m{1.2cm} 
                  >{\centering\arraybackslash}m{1.2cm} 
                  >{\centering\arraybackslash}m{1.2cm} 
                  >{\centering\arraybackslash}m{1.2cm} 
                  >{\centering\arraybackslash}m{1.2cm}} 
    \textbf{Confidence} & 0.9 & 0.8 & 0.7 & 0.6 & 0.5 \\
    \hline
    $b_{\text{pred}}$ &
    \includegraphics[width=1.5cm]{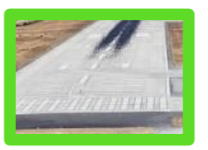} &
    \includegraphics[width=1.5cm]{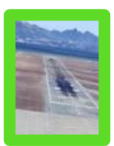} &
    \includegraphics[width=1.5cm]{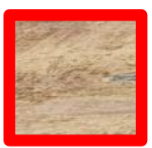} &
    \includegraphics[width=1.5cm]{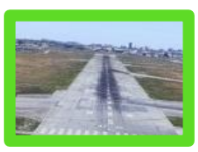} &
    \includegraphics[width=1.5cm]{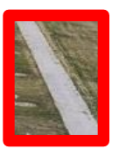} \\
    \hline
    $\text{IoU}(b_{\text{pred}}, b_{\text{gt}})\geq \tau$ & 1 & 1 & 0 & 1 & 0 \\
    \hline
    \textbf{TP} & 1 & 2 & 2 & 3 & 3 \\
    \textbf{FP} & 0 & 0 & 1 & 1 & 2 \\
    \textbf{FN} & 2 & 1 & 1 & 0 & 0 \\
    \hline
    \textbf{Precision = $\frac{TP}{TP+FP}$} & 1 & 1 & $\frac{2}{3}$ & $\frac{3}{4}$ & $\frac{3}{5}$ \\
    \textbf{Recall = $\frac{TP}{TP+FN}$} & $\frac{1}{3}$ & $\frac{2}{3}$ & $\frac{2}{3}$ & 1 & 1 \\
  \end{tabular}

  \caption{Computation of mAP step-by-step on an example from the LARD dataset. The top image shows the visual setup, while the table illustrates how predictions at different confidence levels affect precision and recall.}
  \label{fig:pipeline}
\end{figure}

\section{Related Work}
\label{sec:rw}

\subsection{Uncertainty Quantification for Object Detection}
\label{sec:rw_uq}

Uncertainty quantification (UQ) is essential for object detection, especially in safety-critical applications where erroneous predictions can have significant consequences. Several methods have been proposed to estimate uncertainty, but many involve trade-offs that limit their practical use.

One common approach is Deep Ensembles, which estimates epistemic uncertainty by training multiple models and measuring the variance in their predictions. This technique has shown effectiveness in tasks such as robotic perception \cite{DeepEnsembles}, but its major drawback is the computational cost associated with training several models, making it difficult to scale, particularly in real-time or large-scale applications.

Another popular method is Dropout Sampling, which introduces stochasticity during inference to approximate Bayesian inference. This method has been shown to improve robustness under open-set conditions and enhance the model’s ability to quantify label uncertainty \cite{DropoutSampling}. However, like Deep Ensembles, it requires modifications to the model and training process, which can be challenging for certain applications, especially when real-time performance is crucial.

A more efficient technique, Monte Carlo DropBlock \cite{MCDropBlock}, applies dropout during both training and testing phases, which improves generalization and calibration. While promising, it still relies on stochastic processes like Monte Carlo Dropout, which can lead to increased inference times, limiting its applicability in time-sensitive scenarios.

BayesOD \cite{BayesOD} tackles uncertainty estimation by incorporating Bayesian methods directly into the object detection process. This approach reduces uncertainty error metrics, particularly at the Non-Maximum Suppression (NMS) stage, but it still depends on a Bayesian framework and does not provide statistical guarantees.

While these methods offer valuable contributions to uncertainty estimation, they all come with limitations, especially regarding the need for modifications to the training process, computational complexity, and lack of statistical certificates.

In contrast, Conformal Prediction (CP) offers a more promising alternative by providing performance guarantees without requiring such significant changes to the model training pipeline.

\subsection{Conformal Prediction for Object Detection and Aerospace}
\label{sec:rw_cp}


Conformal Prediction (CP) is a flexible, distribution-free framework that enables the construction of statistically valid prediction sets, under the minimal assumption that data are exchangeable \cite{vovk2022algorithmic, angelopoulos2022conformal}. In its standard form, CP requires a trained model and a separate calibration dataset to compute nonconformity scores—measures of how unusual a new prediction is compared to previously observed outcomes. The result is a prediction set that, with probability approximately $1 - \alpha$, contains the true label. This makes CP particularly attractive for safety-critical applications where quantifying predictive uncertainty is essential.

Extending CP to object detection poses unique challenges, primarily due to the need to conformalize structured outputs like bounding boxes. Recent works have explored ways to adapt CP by introducing pairing strategies between predicted and ground-truth boxes, typically based on Intersection over Union (IoU) thresholds \cite{de2022object, andeol2023confident}. Some approaches apply CP box-wise, computing per-coordinate or per-box nonconformity scores, while others propose more elaborate frameworks such as two-step conformal prediction to handle uncertainty in both localization and classification \cite{timans2024adaptive}. These methods aim to produce conformalized bounding boxes that maintain the desired coverage while remaining practically tight and adaptive.

Robust variants of conformal prediction have also been developed to maintain valid uncertainty guarantees under adversarial perturbations or distribution shifts. For example, Lipschitz-bounded networks have been used to efficiently estimate robust CP sets with strong performance on large-scale datasets, even under adversarial attacks \cite{massena2025efficient}.

In the aerospace domain, CP has been applied across multiple modalities, including surrogate modeling \cite{ducoffe2020high}, satellite image analysis \cite{copley2024uncertain}, and air traffic control systems \cite{ernez2023applying}. 



\section{Single Object Detection for Visual Based Landing}

The Vision Landing System (VLS) pipeline consists of three core stages, each critical for the accurate detection and localization of an aircraft during landing. The full pipeline is captured by Figure~\ref{fig:pipeline}. It is considered and has been revisited by various companies and researchers in the aeronautic field (\cite{daedalean}, \cite{dai2024yomo}):

\begin{itemize}
    \item \textbf{Stage 1 – Object Detection (OD)} \\
    The system receives an input image and identifies the runway within it. This detected region of interest is then cropped to enhance precision in subsequent stages. This step involves training an object detection model to locate and identify runways\\
    
    \item \textbf{Stage 2 – Feature Regression} \\
    From the cropped image, the system predicts runway features—such as its four corners. These features are used to define the runway’s geometry and establish a mapping between 2D image coordinates and their corresponding 3D positions in the real world.

    \item \textbf{Stage 3 – Pose Estimation} \\
    Using the extracted features, the system estimates the aircraft’s pose relative to the runway. This is accomplished using pose estimation techniques, providing real-time aircraft localization during the landing phase.
\end{itemize}

Each stage plays a crucial role in ensuring accurate aircraft localization. In particular, the stage of \textbf{Runway Detection (Stage 1)} forms the foundation of the entire pipeline. Errors at this stage can propagate through the system, degrading both feature extraction and pose estimation accuracy. Therefore, the precision of the initial detection has a direct and significant impact on overall system performance. To ensure reliable corner detection, the cropped image must contain the \textbf{full extent} of the runway and be \textbf{small enough} to avoid inaccuracies introduced by interpolation. Maintaining this balance is essential for preserving an accurate pose estimation.

\begin{figure}[htbp]
\floatconts
  {fig:pipeline}
  {\caption{Overview of the Visual Landing System (VLS) pipeline. The system processes an input image through three stages: (1) Object Detection to identify and crop the runway region, (2) Feature Regression to extract key geometric features such as corners, and (3) Pose Estimation to compute the aircraft’s position relative to the runway using the extracted features. Each stage builds upon the previous one, with accurate detection being critical to the reliability of the final pose estimation.}}
  {\includegraphics[width=1.\linewidth]{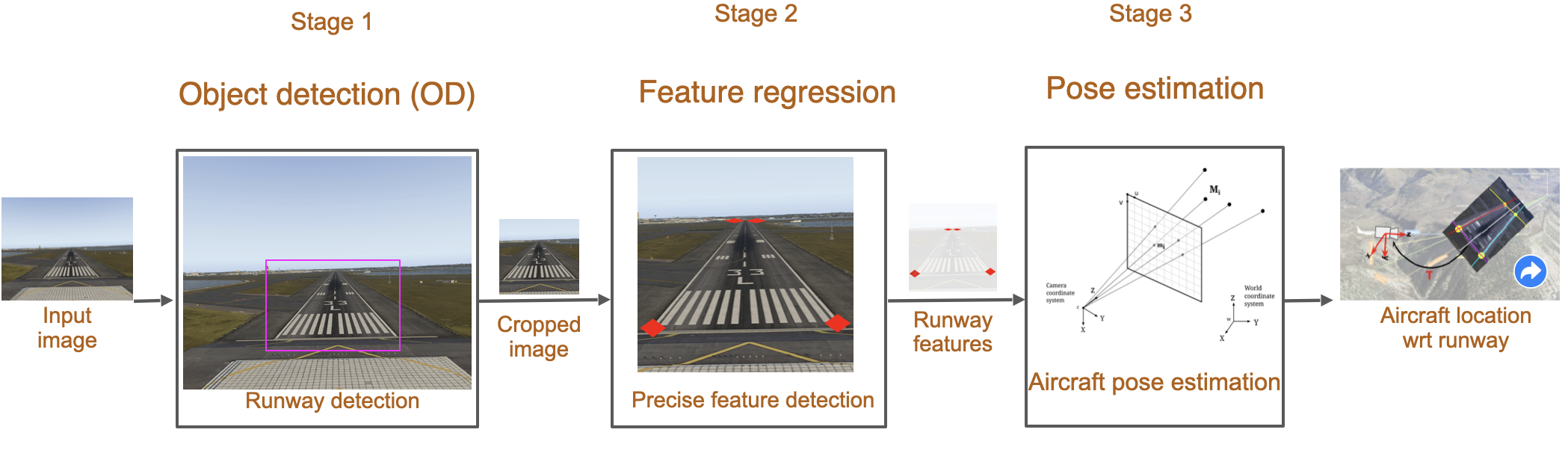}}
\end{figure}

\subsection{On Robust Detection Metrics: IoU vs. IoA}

\begin{figure}[H]
  \centering
  \includegraphics[height=4cm]{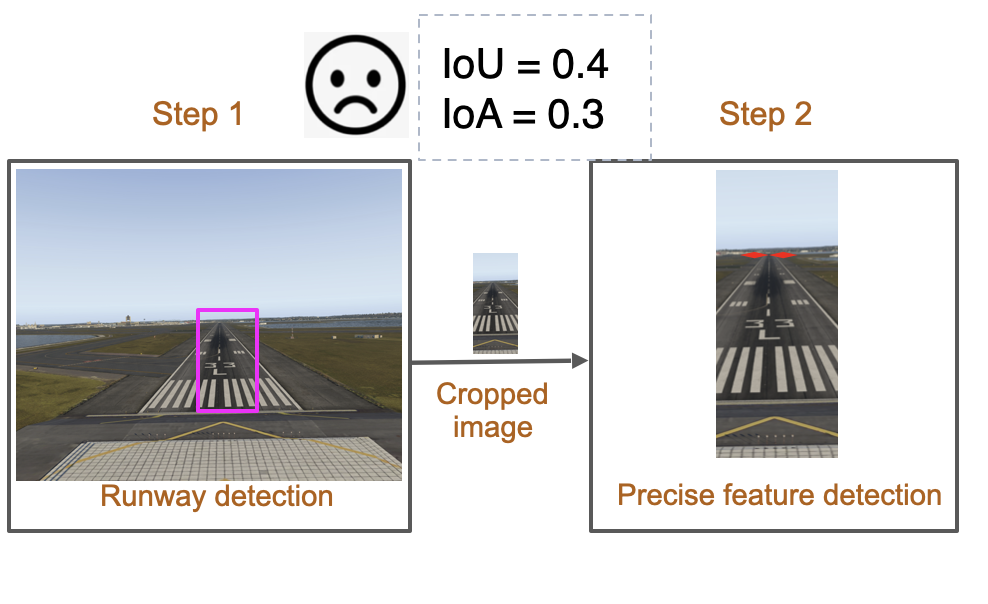}
  \hspace{0.02\linewidth}
  \includegraphics[height=4cm]{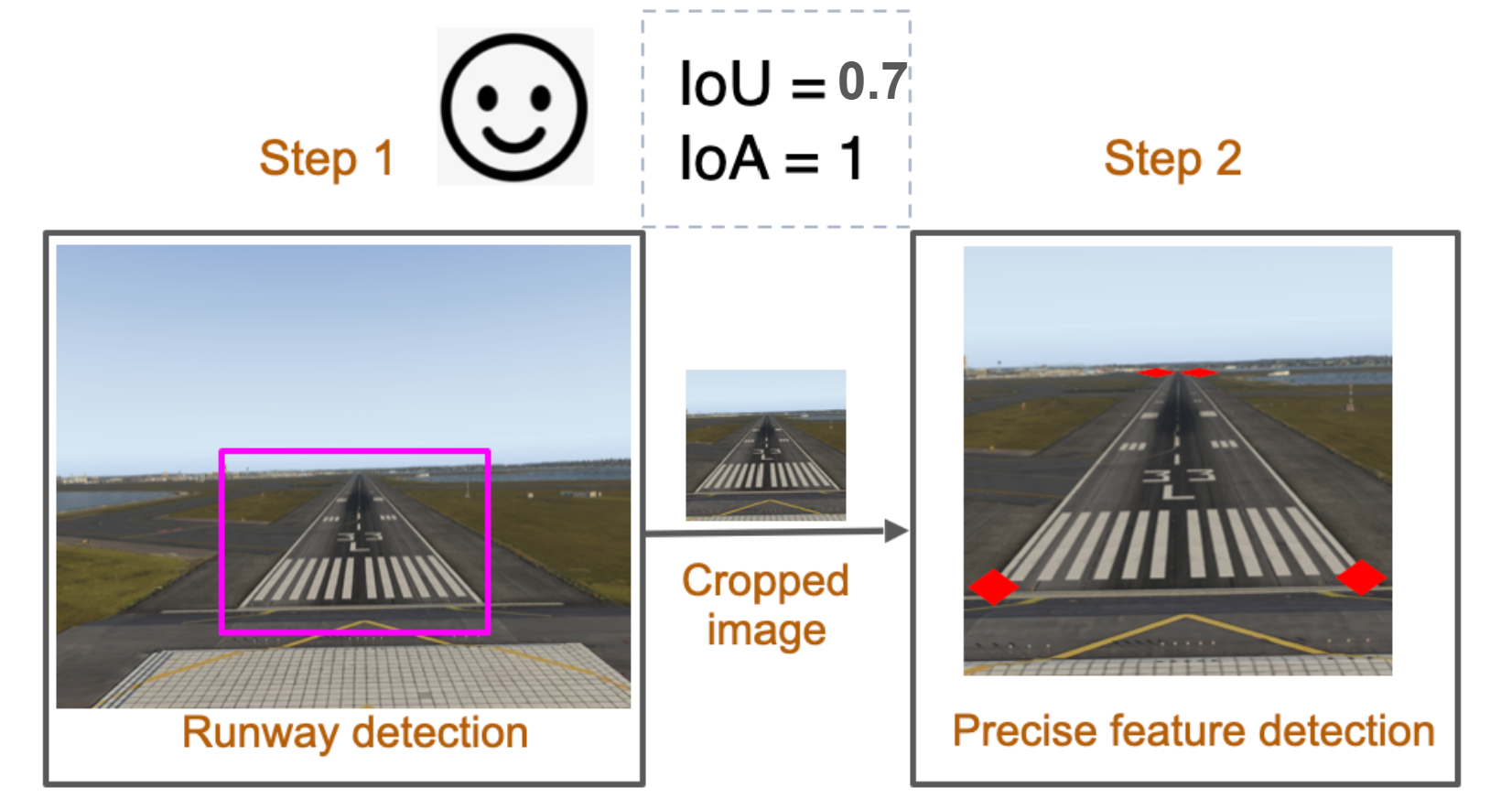}
  \caption{Pipeline examples illustrating different IoU and IoA outcomes.}
  \label{fig:iou_ioa}
\end{figure}

Given the importance of accurately identifying the runway region in Stage 1, it is crucial to evaluate object detection performance using metrics that reflect both spatial precision and complete coverage. While Intersection over Union (IoU) is the conventional choice for object detection evaluation, it does not fully align with the dual objectives of our system illustrated in Figure~\ref{fig:iou_ioa}.

\begin{enumerate}
    \item The predicted bounding box should \textit{fully enclose} the ground truth box to ensure reliable feature regression and pose estimation. This is particularly critical, as pose estimation becomes unreliable with fewer than three visible runway corners.
    \item The predicted bounding box should also be \textit{accurately aligned} with the ground truth to avoid interpolation error when estimating key point features.
\end{enumerate}


IoU is well-suited for assessing criterion (2), as it captures the degree of overlap between the predicted and ground truth boxes. However, it does not guarantee full coverage required by criterion (1). To address this limitation, we incorporate an additional metric: \textbf{Intersection over Area (IoA)}, defined as the intersection area divided by the area of the ground truth bounding box.

IoA has been previously employed in the conformalized object detection literature (e.g., \cite{de2022object}) due to its \textit{monotonicity}. This property implies that increasing the size of the predicted box to better enclose the ground truth results necessarily in a higher IoA score. In contrast, IoU may decrease in such scenarios if the added area extends beyond the ground truth, making IoA more compatible with Conformal Prediction (CP) frameworks that require such predictable behavior.

We define a detection as \textit{robust} when both spatial accuracy and full coverage are satisfied:

\[
\text{Robust Detection} \iff \text{IoU}(\textbf{b}_{\text{pred}}, \textbf{b}_{\text{gt}}) \geq t \quad \text{and} \quad \text{IoA}(\textbf{b}_{\text{pred}}, \textbf{b}_{\text{gt}}) = 1,
\]

where $\textbf{b}_{\text{pred}}$ and
$\textbf{b}_{\text{gt}}$ denote the 
associated predicted and ground truth 
bounding boxes, respectively.

\subsection{Conformal Average Precision (C-AP)}

In Conformal Single Object Detection, where we focus on detecting a single class—specifically the runway—we introduce Conformal Average Precision (C-AP). 
Our method is straightforwardly applicable to to multiple object detection tasks, where the C-mAP metric can evaluate performance across multiple classes. Our approach proceeds in two steps:
\begin{enumerate}
    \item We apply a Conformal Prediction (CP) framework to ensure that the IoA equals 1 for the majority of true positive predictions.
    \item We introduce a new metric, \textit{Conformal Average Precision (C-AP)}, to quantify the trade-off between constraints (1) and (2) using a modified version of Average Precision (AP).
\end{enumerate}

First we adopt a conformal prediction (CP) framework that leverages the discrepancy between predicted and ground-truth bounding box coordinates. Specifically, we follow the method proposed by \citet{de2022object}, where the nonconformity score is defined based on coordinate-wise differences between predicted and groundtruth bounding boxes.

Let each ground-truth bounding box be indexed by $k = 1, \dots, N_{gt}$, regardless of the image it appears in. We denote the ground-truth coordinates of the $k$-th box as $\mathbf{b}^{gt}_k = (x^k_{\min}, y^k_{\min}, x^k_{\max}, y^k_{\max})$, and its predicted counterpart as $\mathbf{\hat{b}}_k = (\hat{x}^k_{\min}, \hat{y}^k_{\min}, \hat{x}^k_{\max}, \hat{y}^k_{\max})$. The \textbf{additive } nonconformity measure $\mathbf{r}_k$ is then defined as the vector:
\[
\mathbf{r}^a_k = \left( \hat{x}^k_{\min} - x^k_{\min}, \; \hat{y}^k_{\min} - y^k_{\min}, \; x^k_{\max} - \hat{x}^k_{\max}, \; y^k_{\max} - \hat{y}^k_{\max} \right).
\]

while the \textbf{multiplicative } nonconformity measure $\mathbf{r}_k$ is defined using the width and height of the predicted bounding box $\hat{w} = \hat{x}_{max} - \hat{x}_{min}$, $\hat{h} = \hat{y}_{max} - \hat{y}_{min}$

\[
\mathbf{r}^m_k = \left( \frac{\hat{x}^k_{\min} - x^k_{\min}}{\hat{w}}, \; \frac{\hat{y}^k_{\min} - y^k_{\min}}{\hat{h}}, \; \frac{x^k_{\max} - \hat{x}^k_{\max}}{\hat{w}}, \; \frac{y^k_{\max} - \hat{y}^k_{\max}}{\hat{h}}\right).
\]

Similarly as previous approaches, we use the Hungarian matching algorithm during the conformalization of predicted boxes.





Once these pairings are established, we compute nonconformity scores for conformal prediction. Since our bounding box predictions are represented as 4-dimensional vectors (one value for each side of the box), we obtain four separate distributions of nonconformity scores—unlike the scalar scores used in standard conformal prediction. 
To maintain valid statistical coverage guarantees across all dimensions, we apply a Bonferroni correction, as originally proposed in~\cite{de2022object}.
That is, we build quantiles for each of the 4 non-conformity component, at level $\frac{\alpha}{4}$, resulting in
\[
    \hat{q}(j) = q_{\left\lceil (1 - \frac{\alpha}{4})(n + 1) \right\rceil / n}(\{\mathbf{r}_k(j) : k \in \{1,\dots,n\}\}),
\]
where $q_{\left\lceil (1 - \frac{\alpha}{4})(n + 1) \right\rceil / n}$ is the \emph{corrected} quantile function at level $\frac{\alpha}{4}$, and  $r_k(j)$ as well as $\hat{q}(j)$ are respectively the $j$-th component of the vector $r_k$ and $\hat{q}$. 

The resulting quantiles from each score distribution are then used to adjust predictions at inference. For a predicted bounding box $\mathbf{\hat{b}}_k$, we construct the \textit{conformal} bounding box $\mathbf{\hat{b}}_k^\mathrm{conf}$ using the quantiles $\mathbf{\hat{q}}$ following additive and multiplicative corrections of \cite{de2022object}. The conformal bounding box is theoretically guaranteed to contain the true bounding box with a user-specified probability at least $1-\alpha$.
More formally, we have

\[
    \mathbb{P}\left(\mathbf{b}_{n+1}^{gt} \subseteq \mathbf{\hat{b}}_{n+1}^\mathrm{conf}\right)\geq 1-\alpha,
\]
where the $b\subseteq b^\prime$ corresponds to the box inclusion (i.e., holding true if $x_{min}\geq x_{min}^\prime,\ y_{min}\geq y_{min}^\prime,\ x_{max} \leq x_{max}^\prime,\ y_{max}\leq y_{max}^\prime$). 

\begin{figure}[H]
  \centering
  \includegraphics[width=\linewidth]{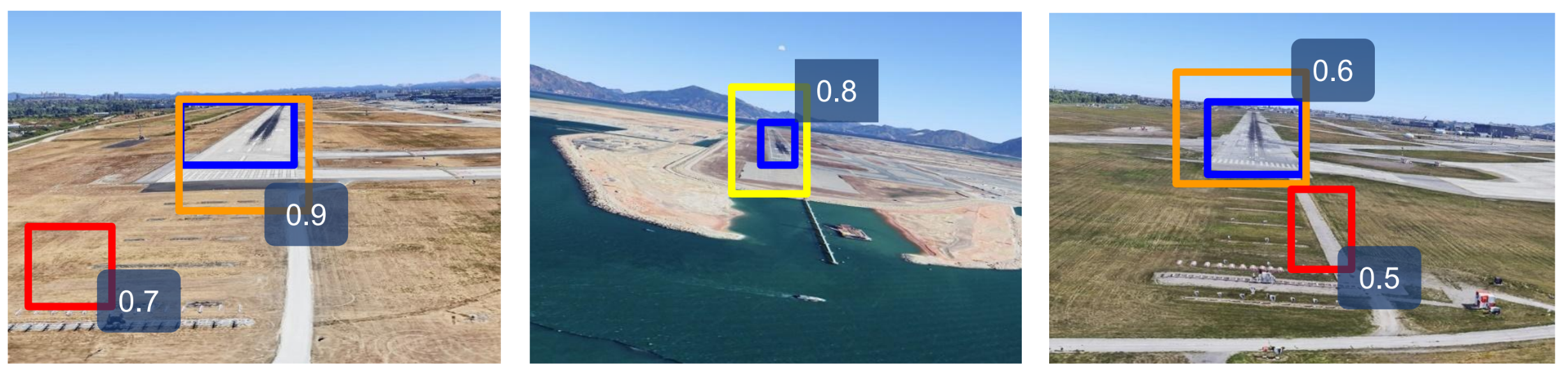}

  \vspace{0.5em} 

  \begin{tabular}{>{\centering\arraybackslash}m{3.5cm} 
                  >{\centering\arraybackslash}m{1.2cm} 
                  >{\centering\arraybackslash}m{1.2cm} 
                  >{\centering\arraybackslash}m{1.2cm} 
                  >{\centering\arraybackslash}m{1.2cm} 
                  >{\centering\arraybackslash}m{1.2cm}} 
    \textbf{Confidence} & 0.9 & 0.8 & 0.7 & 0.6 & 0.5 \\
    \hline
    $b_{\text{pred}}$ &
    \includegraphics[width=1.5cm]{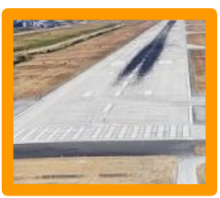} &
    \includegraphics[width=1.5cm]{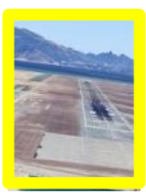} &
    \includegraphics[width=1.5cm]{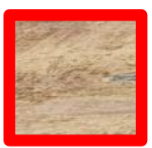} &
    \includegraphics[width=1.5cm]{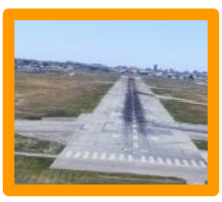} &
    \includegraphics[width=1.5cm]{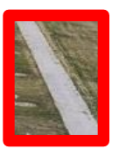} \\
    \hline
    $\text{IoU}(b_{\text{pred}}, b_{\text{gt}})\geq \tau$ & 1 & 0 & 0 & 1 & 0 \\
    \hline
    $\text{IoA}(b_{\text{pred}}, b_{\text{gt}})=1$ & 1 & 1 & 0 & 1 & 0 \\
    \hline
    \textbf{TP} & 1 & 1 & 1 & 2 & 2 \\
    \textbf{FP} & 0 & 1 & 2 & 2 & 3 \\
    \textbf{FN} & 2 & 2 & 2 & 1 & 1 \\
    \hline
    \textbf{Precision = $\frac{TP}{TP+FP}$} & 1 & $\frac{1}{2}$ & $\frac{1}{3}$ & $\frac{1}{2}$ & $\frac{2}{5}$ \\
    \textbf{Recall = $\frac{TP}{TP+FN}$} & $\frac{1}{3}$ & $\frac{1}{3}$ & $\frac{1}{3}$ & $\frac{2}{3}$ & $\frac{2}{3}$ \\
  \end{tabular}

  \caption{Computation of \textbf{C-(m)AP} step-by-step on an example from the LARD dataset. The top image shows the visual setup, while the table illustrates how predictions at different confidence levels affect precision and recall.}
  \label{fig:c_map}
\end{figure}

In the traditional Average Precision (AP) framework, a predicted box $\mathbf{\hat{b}}_k$ is considered a true positive if it satisfies $\text{IoU}(\mathbf{\hat{b}}_k, \mathbf{b}^{\text{gt}}_k) \geq t$. In our conformal AP (C-AP) formulation, we strengthen this condition, as illustrated in Figure~\ref{fig:c_map}:

\[
\text{C-AP Match} \iff \text{IoU}(\mathbf{\hat{b}}_k, \mathbf{b}^{gt}_k) \geq t \quad \text{and} \quad \text{IoA}(\mathbf{\hat{b}}_k, \mathbf{b}^{gt}_k) = 1.
\]

This stricter requirement ensures that a prediction is only counted as a true positive if it not only sufficiently overlaps the ground truth but also fully contains it—better aligning detection evaluation with the downstream requirements of accurate pose estimation.

\section{Experiments}

Our experiments seek to demonstrate the utility of conformal prediction approaches to
quantify localization uncertainty which exist in object detection models.

\textbf{Dataset}: the LARD dataset \cite{ducoffe2023lard} comprises high-quality aerial images of runway during approach and landing phases. 
We split the dataset into train, validation and test. The mAP of the models are evaluated on the validation set while the experiments on conformal prediction are done on the the remaining 20\% of the test set. The number of image per split is described in Table~\ref{tab:dataset_composition}. The LARD dataset contains both synthetic images and real images but the real images are only kept in the test set.
The current version of the dataset contains only one ground truth runway annotation per image. However, our method is already designed to generalize to multiple ground truth instances if they become available in future versions.

\begin{table}[h]
\centering
\begin{tabular}{llr}
\textbf{Set} & \textbf{Type} & \textbf{Images} \\
\midrule
Train & Synthetic & 11,546 \\
Validation & Synthetic & 2,886 \\
Test & Real+Synth & 2,315 \\
\end{tabular}
\caption{Dataset composition by set, type, and number of images.}
\label{tab:dataset_composition}
\end{table}

\subsection{Models Description and Evaluation}

In this section, we describe the object detection models employed for the runway detection task and evaluate their performance on aerial imagery. We benchmark two state-of-the-art, pre-trained models from the YOLO (You Only Look Once) family: YOLOv5-small \cite{jocher2020ultralytics} and YOLOv6-small \cite{li2022yolov6}, chosen for their balance between detection accuracy and computational efficiency—an essential requirement for embedded vision-based landing systems.

\subsubsection{Model Configurations and Training}
Both models were initialized from pre-trained weights available through their respective official repositories and fine-tuned on our runway detection dataset using standard training protocols.
YOLOv5-small was trained for 100 epochs with a batch size of 16 and an input resolution of 640×640 pixels. This configuration aligns with common practices in the YOLO community and was found sufficient for convergence.
YOLOv6-small was trained for 97 epochs, using a batch size of 32 and the same input resolution of 640×640. Training was conducted using the official YOLOv6 repository and default optimization settings.

Both training procedures incorporated extensive data augmentation, which is critical for generalization on aerial imagery with varying lighting, weather, and perspectives, using the default configuration.

\subsubsection{Model Evaluation}
The models were evaluated using standard object detection metrics: mean Average Precision at IoU 0.5 (mAP@0.5) and mean Average Precision across IoU thresholds from 0.5 to 0.95 (mAP@0.5:0.95). These metrics offer insights into both loose and strict detection accuracies. Additionally, we report GFLOPs (Giga Floating Point Operations) as a proxy for computational efficiency—an important constraint for real-time, onboard inference.

As shown in Table~\ref{tab:model_comparison}, both models achieve near-perfect performance on the runway detection dataset. YOLOv5 achieves slightly higher precision, especially at tighter IoU thresholds, while YOLOv6 presents a more computationally demanding architecture, as indicated by its GFLOPs. These results provide a strong baseline for applying conformal prediction methods, where bounding box uncertainty will be layered atop already accurate detections. Figure~\ref{fig:comparison_iou} displays a scatter plot comparing the performance of YOLOv5 and YOLOv6. The x-axis represents the Intersection over Union (IoU) between YOLOv5’s predictions and the ground-truth boxes, while the y-axis shows the IoU for YOLOv6. The distribution of points suggests no clear correlation between the IoU values of the two models. This indicates that the models behave differently across images, often producing divergent predictions for the same inputs.

\begin{figure}[htbp]
\floatconts
  {fig:comparison_iou}
  {\caption{Scatter plot comparing the IoU of YOLOv5 (x-axis) and YOLOv6 (y-axis) with respect to the ground-truth boxes. Each point represents a single image. The lack of a clear correlation suggests that the two models produce different predictions on the same inputs.}
}
  {\includegraphics[width=0.5\linewidth]{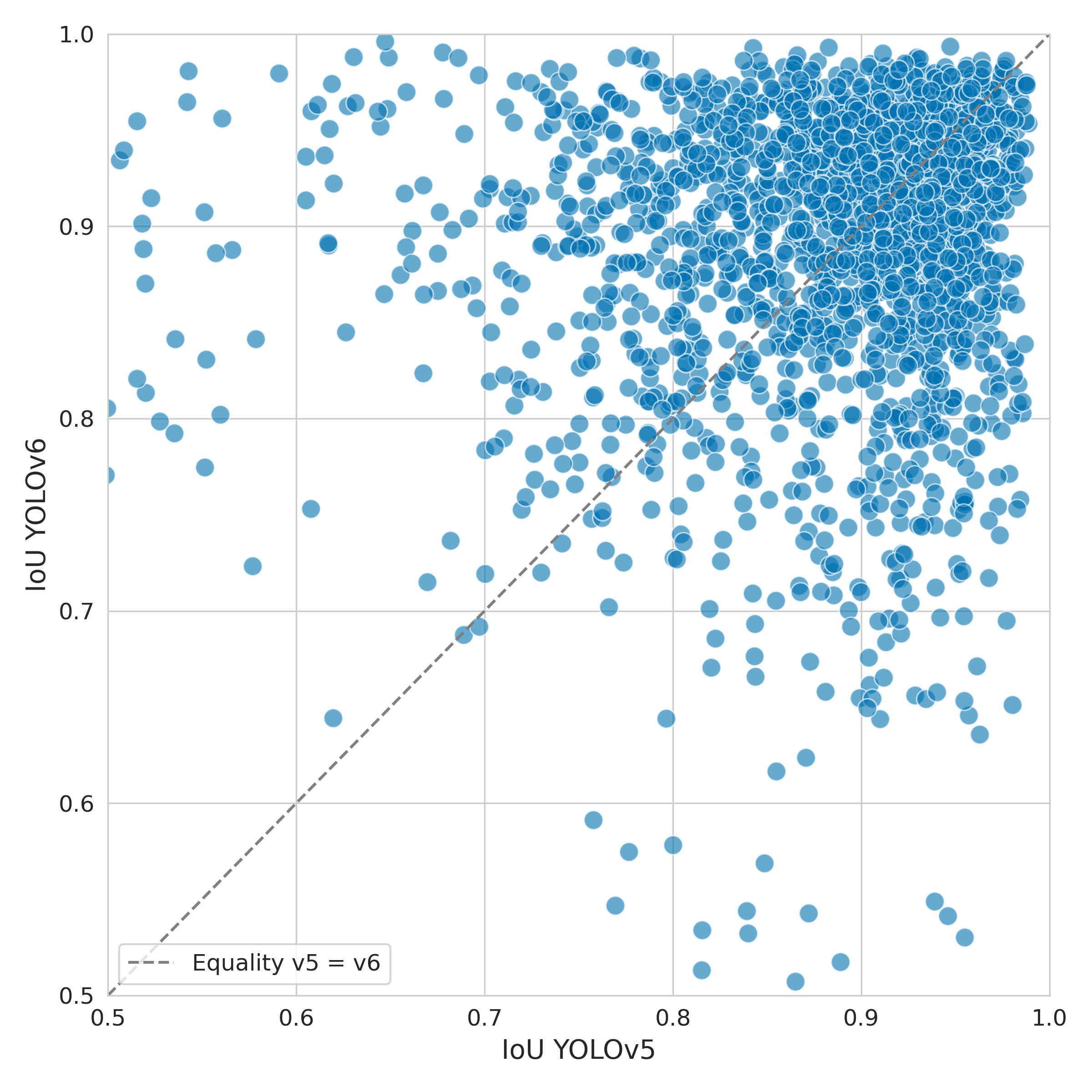}}
\end{figure}

\begin{table}[ht]
\centering
\caption{Comparison of pre-trained object detection models on runway detection task evaluated on the validation dataset.}
\label{tab:model_comparison}
\begin{tabular}{lcc}
\textbf{Metric} & \textbf{YOLOv5 (pre-trained)} & \textbf{YOLOv6 (pre-trained)} \\
\midrule
mAP@0.5         & 0.995                         & 0.9901 \\
mAP@0.5:0.95    & 0.9712                        & 0.9413 \\
GFLOPs          & 15.8                          & 45.3 \\
\end{tabular}
\end{table}

\subsection{Conformal Prediction for Object Localization}

We evaluate the effect of conformalization on object detection models by analyzing the coverage and bounding box area after applying conformal prediction  with error level $\alpha = 0.3$. We use the \texttt{PUNCC} library \cite{mendil2023puncc}. We define respectively by c-YOLOv5-${a,m}$ and c-YOLOv6-${a,m}$ the conformalized versions of the original YOLOv5 and YOLOv6 models, with \textit{a} denoting the additive penalty and \textit{m} the multiplicative penalty used to enlarge predicted bounding boxes. 
Table~\ref{tab:yolo_metrics} reports the proportion of predicted boxes that cover their associated ground truth (coverage), along with the average area and variance of the conformalized predicted boxes.

Our results show that the empirical coverage often exceeds the target level $(1 - \alpha)$. This conservativeness is likely a result of the Bonferroni correction, which tends to inflate the quantiles to ensure joint coverage over multiple dimensions. Consequently, depending on the baseline accuracy of the detection model, enforcing stricter coverage (lower $\alpha$) or using more conservative nonconformity scores will generally produce larger bounding boxes to maintain the desired guarantees.
Among the evaluated models, c-YOLOv5-a achieves the highest coverage at 77.06\%, followed by c-YOLOv5-m (75.88\%), c-YOLOv6-a (75.73\%), and c-YOLOv6-m (73.93\%). This indicates that the additive penalty tends to provide slightly better coverage than the multiplicative one, particularly for YOLOv5. 

 Our next study will compare the Conformal Average precision of all models. We evaluate the performances of standard YOLO models (YOLOv5 and YOLOv6) and their conformalized variants (c-YOLOv5 with additive penalty, c-YOLOv5 with multiplicative penalty, c-YOLOv6 with additive penalty, and c-YOLOv6 with multiplicative penalty) across three metrics: standard mean Average Precision (mAP), Conformal mean Average Precision (C-mAP, naturally extended from C-mAP), and C-mAP@50@80:100, a variant measuring performance with IoU threshold set to 0.5 and across varying IoA thresholds from 0.8 to 1.0. Metrics are computed on the Test Set.

The results, reported in Table~\ref{tab:c_map_results}, reveal clear trends:

\begin{table}[H]
\centering
\begin{tabular}{lccc}
\textbf{Model} & \textbf{mAP}& \textbf{C-mAP}&\textbf{C-mAP@50@80:100} \\
\midrule
YOLOv5 & 96.88 & 0.77 & 46.92 \\
c-YOLOv5-a & 92.67 & \textbf{56.86} & 80.73\\
c-YOLOv5-m & 96.17 & 55.84& \textbf{82.18}\\
YOLOv6 & \textbf{98.13} & 1.31& 51.94 \\
c-YOLOv6-a & 95.09 & 55.75 & 81.86 \\
c-YOLOv6-m & 96.71 & 52.71 & 81.93 \\
\end{tabular}
\caption{Evaluation of YOLOv5 and YOLOv6 and their conformalized version on the Global Test Dataset}
\label{tab:c_map_results}
\end{table}

First, standard YOLO models achieve very high mAP scores (96.88\% for YOLOv5 and 98.13\% for YOLOv6), confirming their excellent performance under traditional detection criteria based solely on Intersection over Union (IoU). However, their C-mAP scores are extremely low (0.77\% and 1.31\%, respectively), indicating that very few predictions fully contain the ground truth boxes as required by the stricter conformal matching condition. This gap highlights that conventional detectors, while spatially accurate in terms of overlap, often fail to completely encapsulate objects---a critical property for downstream tasks such as precise pose estimation.

In contrast, the conformalized models significantly improve C-mAP scores, achieving between 52.71\% and 56.86\%, while maintaining high standard mAP values (above 92\% in all cases). This demonstrates that the conformalization process, whether via an additive or a multiplicative penalty, effectively adapts the models to produce predictions that not only overlap but also fully contain ground truths, with only a minor compromise on traditional detection performance. The intermediate metric, C-mAP@50@80:100, further supports this observation. While vanilla YOLO models achieve moderate scores (46.92\% and 51.94\%), their conformalized counterparts reach substantially higher results (approximately 81--82\%). This suggests that conformalized models remain robust across a range of IoA thresholds, consistently maintaining high-quality detections even as the containment requirement becomes more stringent.

Finally, between the additive and multiplicative penalty variants, the multiplicative models generally achieve a slightly better balance, preserving higher standard mAP scores while maintaining competitive C-mAP and C-mAP@50@80:100 performances.

In summary, these results demonstrate the effectiveness of conformalization in adapting object detectors to stricter, application-driven evaluation criteria. Furthermore, they highlight the relevance of our proposed Conformal mAP (C-mAP) metric, which simultaneously encapsulates both precision (through IoU requirements) and full coverage (through IoA constraints), offering a more holistic and practical assessment of detection performance for downstream applications where spatial guarantees are crucial.

In order to quantify the impact of conformalization over our base predictor, we compute the margin and expansion of the conformalized boxes as:  

\[
\textrm{Margin}(d): \quad \frac{1}{n} \sum_{i=1}^{n} \left| \mathbf{\hat{b}}^\mathrm{conf}_{i, d} - \mathbf{\hat{b}}_{i,d} \right|,
\]

where $\mathbf{\hat{b}}_{i,d}^\mathrm{conf}$ and $\mathbf{\hat{b}}_{i,d}$ correspond respectively to the $d$-th index of the $i$-th conformalized and predicted box, and $n$ the number of boxes.

The margins in Table \ref{tab:pred_margins} show how much
the predicted boxes were expanded to form the conformalized boxes, with higher values indicating significant expansion and smaller values suggesting minimal 
adjustments. Balanced margins reflect symmetrical corrections.
c-YOLOv5-a and c-YOLOv6-a have smaller margins, with c-YOLOv6-a 
exhibiting the least expansion across all directions. This suggests 
high model confidence and minimal adjustments, indicating better 
prediction accuracy. In contrast, c-YOLOv5-m requires larger 
adjustments, especially in the left (18.17) and right (12.04) 
directions, implying lower model confidence and more substantial 
corrections. The Total Average column in Table \ref{tab:pred_margins} 
provides an overall measure of expansion across all directions. A lower 
value indicates more precise predictions, while a higher value suggests 
more adjustments. c-YOLOv6-a has the lowest total average margin 
(7.03), reflecting better accuracy, while c-YOLOv5-m has the highest total average margin (12.18), indicating the model requires larger 
corrections.

\begin{table}[h]
\centering
\begin{tabular}{lccccc}
\textbf{Model} & \textbf{Left} & \textbf{Top} & \textbf{Right} & \textbf{Bottom} & \textbf{Total Average} \\
\midrule
c-YOLOv5-a & 11.76 & 5.79 & 8.93 & 8.54 & 8.76 \\
c-YOLOv5-m & 18.17 & 7.85 & 12.04 & 10.65 & 12.18 \\
\textbf{c-YOLOv6-a}  & \textbf{10.39}  & \textbf{3.47}  & \textbf{7.76}  & \textbf{6.52}  & \textbf{7.03}  \\
c-YOLOv6-m & 15.58 & 4.92 & 11.75 & 7.32 & 9.89 \\
\end{tabular}
\caption{Average predictive margins (in pixels) per side for each conformalized YOLO model.}
\label{tab:pred_margins}
\end{table}

Furthermore, we calculate the \text{Stretch} metric (e.g., \cite{andeol2023confident}), which quantifies the expansion ratio of the conformalized box area relative to the predicted box area. This metric provides insight into the confidence of the model’s predictions, helping to determine whether the model is making small, precise adjustments or compensating for under-covering predictions with larger corrections. The formula is as follows:

\[
\text{Stretch} = \frac{1}{n} \sum_{i=1}^{n} \sqrt{\frac{\text{Area}(C_i)}{\text{Area}(P_i)}}
\]

Where \( \text{Area}(C_i) \) is the area of the conformalized box \( C_i \), and \( \text{Area}(P_i) \) is the area of the predicted box \( P_i \).

A \text{Stretch} value near 1 indicates minimal expansion of the predicted box, suggesting high model confidence with only slight adjustments needed. In contrast, a significantly higher stretch value points to a larger expansion of the predicted box, typically indicating greater uncertainty in the prediction. This suggests that the model compensates for an under-covering prediction by expanding the box to ensure the entire object is captured, reflecting the model's uncertainty about the object boundaries. 

In the results (Table \ref{tab:yolo_metrics}), c-YOLOv6-m shows the lowest Stretch value (1.11), indicating that the model's predictions are the closest to the conformalized boxes, with minimal adjustment needed. c-YOLOv5-a, on the other hand, has a Stretch value of 1.16, indicating slightly more expansion, which suggests a bit more uncertainty or adjustments. Regarding Variance and the mean of the square root of box area, we observe that c-YOLOv6-a has the smallest variance (156.60) and mean area (188.76), suggesting that it exhibits more stable and consistent box areas across predictions. In contrast, c-YOLOv5-m shows a higher variance (177.45) and area (200.42), indicating that the predicted boxes are more spread out and less consistent, reflecting greater variability in the model's predictions.

\begin{table}[h]
\centering
\begin{tabular}{lccc}
\textbf{Model} & \textbf{Stretch} & $\sqrt{\textbf{Box Area}}$ & \textbf{Coverage} \\
\midrule
c-YOLOv5-a & 1.16 & $ 194.92 \pm 156.74$ & \textbf{77.06}\\
c-YOLOv5-m & 1.13 & $200.42 \pm 177.45$ &  75.88\\
c-YOLOv6-a & 1.13 & $\textbf{188.76} \pm \textbf{156.60}$ & 75.73\\
c-YOLOv6-m & \textbf{1.11} &  $193.05 \pm 173.47$ & 73.93\\

\end{tabular}
\caption{Evaluation of YOLOv5 and YOLOv6 on the test dataset with different conformalized box penalty, with associated coverage, squared root of box area and stretch for $\alpha = 0.3$.}
\label{tab:yolo_metrics}
\end{table}

To further understand the behavior of the models, we now turn to the visual analysis of the Intersection over Union (IoU) and Intersection over Area (IoA) distributions. Figures~\ref{fig:comparison_ioa_iou_conformal} and~\ref{fig:comparison_iou_iou_conformal} provide deeper insights into how the conformalized YOLO models compare to the standard YOLO models, specifically in terms of their precision, coverage, and the impact of conformalization on false positives and false negatives.

\begin{figure}[h]
\floatconts
  {fig:comparison_ioa_ioa_conformal} 
  {\caption{Scatter plots comparing the Intersection over Area (IoA) of YOLOv5 and YOLOv6 predictions (x-axis) with their conformalized versions. Left: multiplicative penalty (boxes expanded by a percentage). Right: additive penalty (boxes enlarged by a fixed number of pixels).
}} 
  { 
    \includegraphics[width=0.49\linewidth]{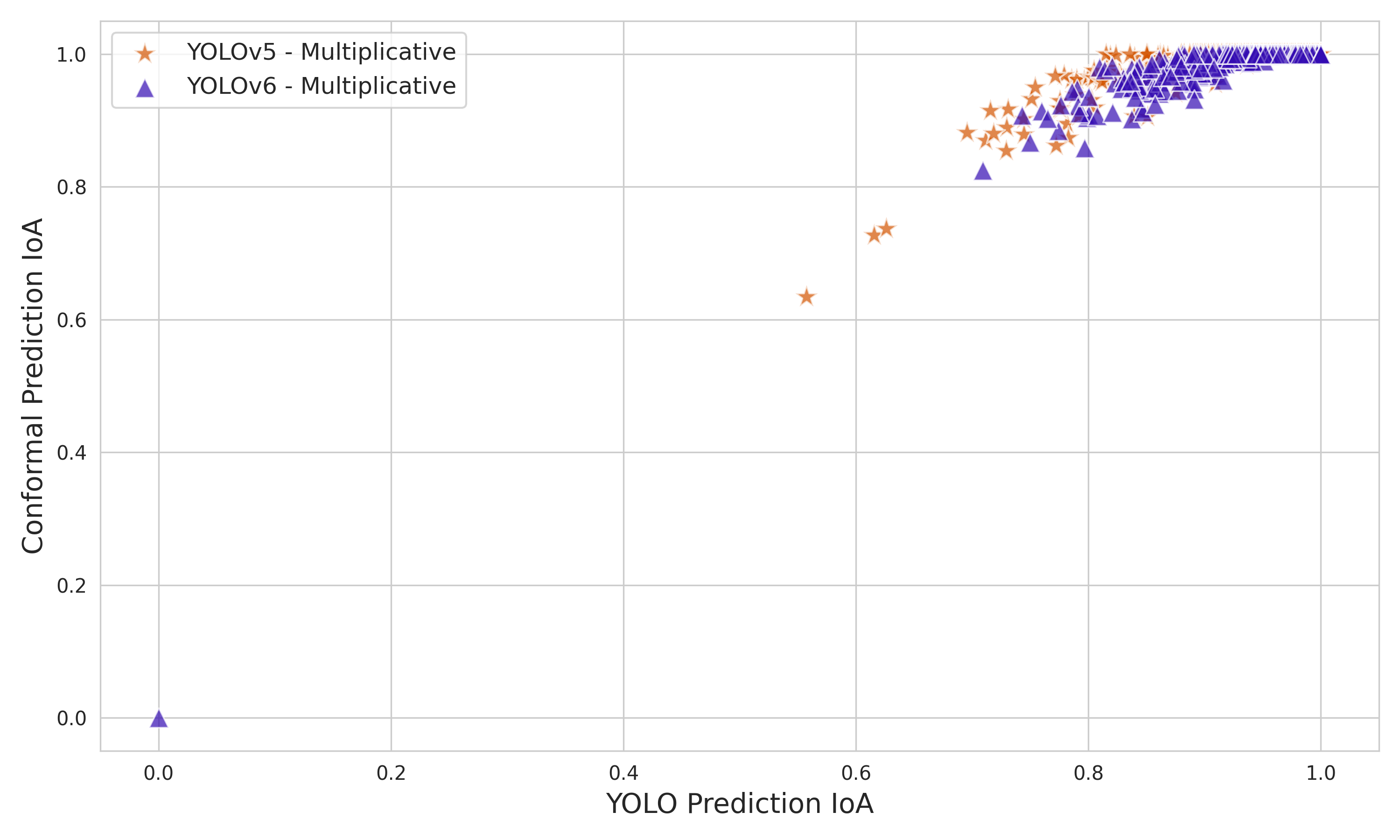}
    \includegraphics[width=0.49\linewidth]{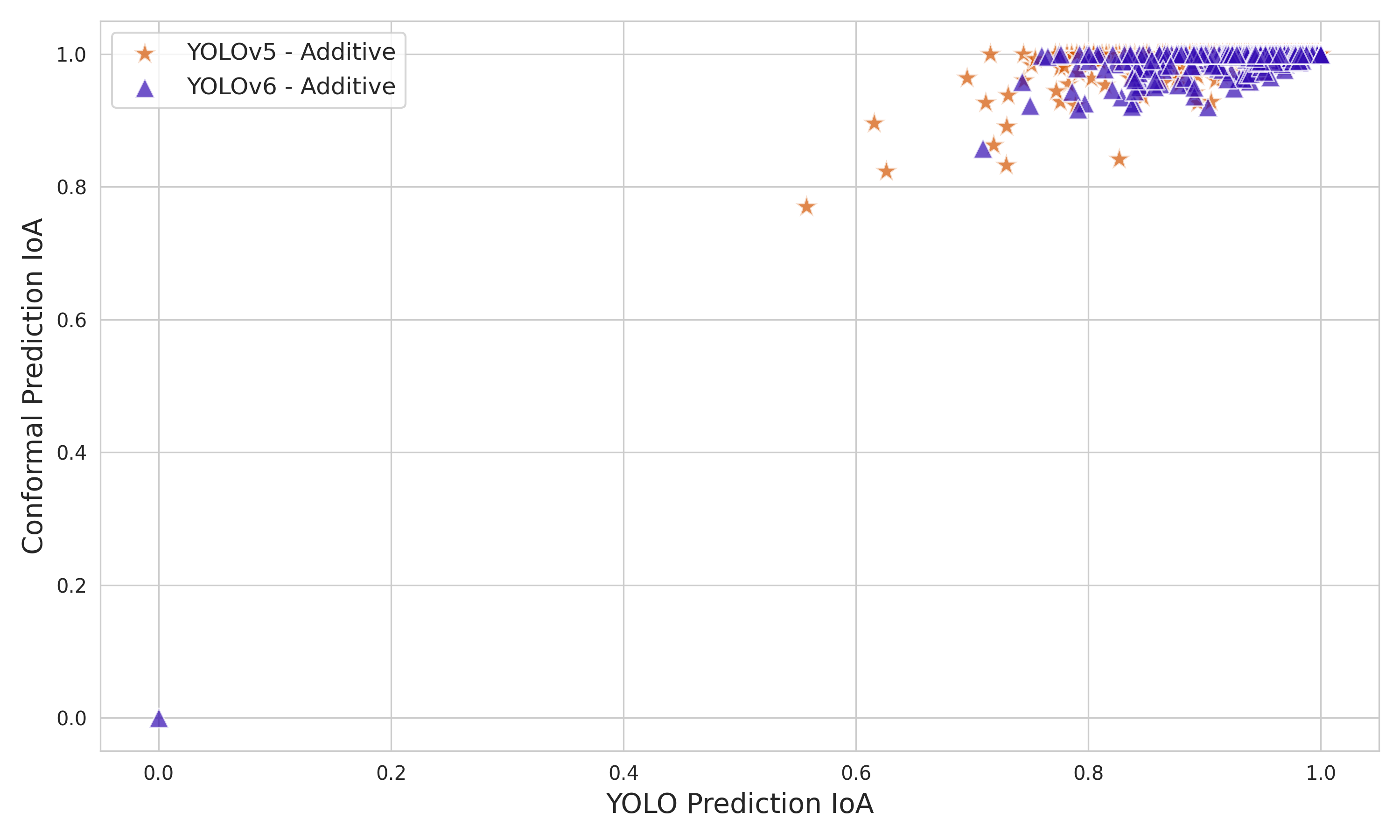}
  }
\end{figure}

\begin{figure}[h]
\floatconts
  {fig:comparison_ioa_iou_conformal} 
  {\caption{Scatter plots comparing the Intersection over Union (IoU) of YOLO'S predictions (x-axis) with the Intersection over Area (IoA) of their conformalized versions. Left: multiplicative penalty. Right: additive penalty.
}} 
  { 
    \includegraphics[width=0.48\linewidth]{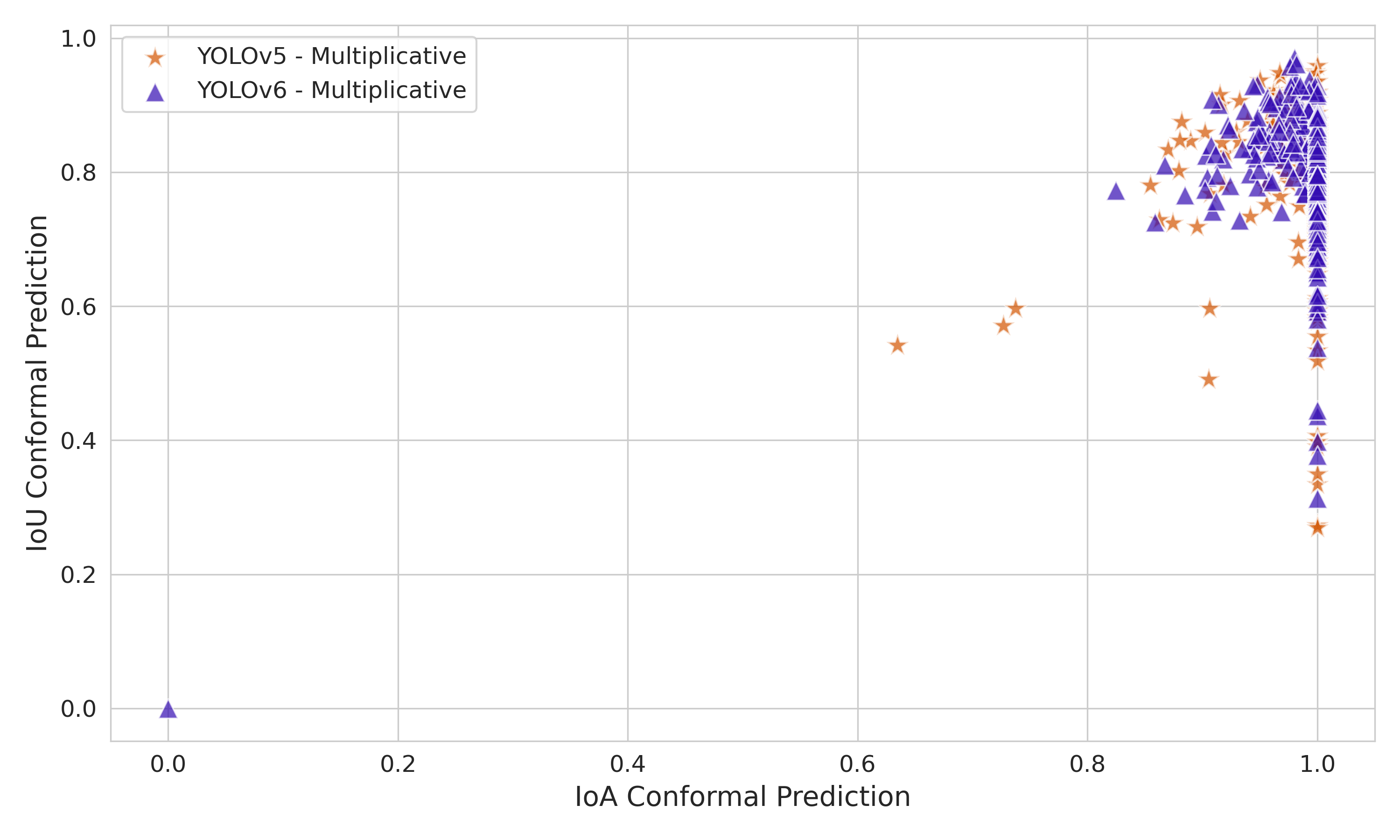}
    \includegraphics[width=0.48\linewidth]{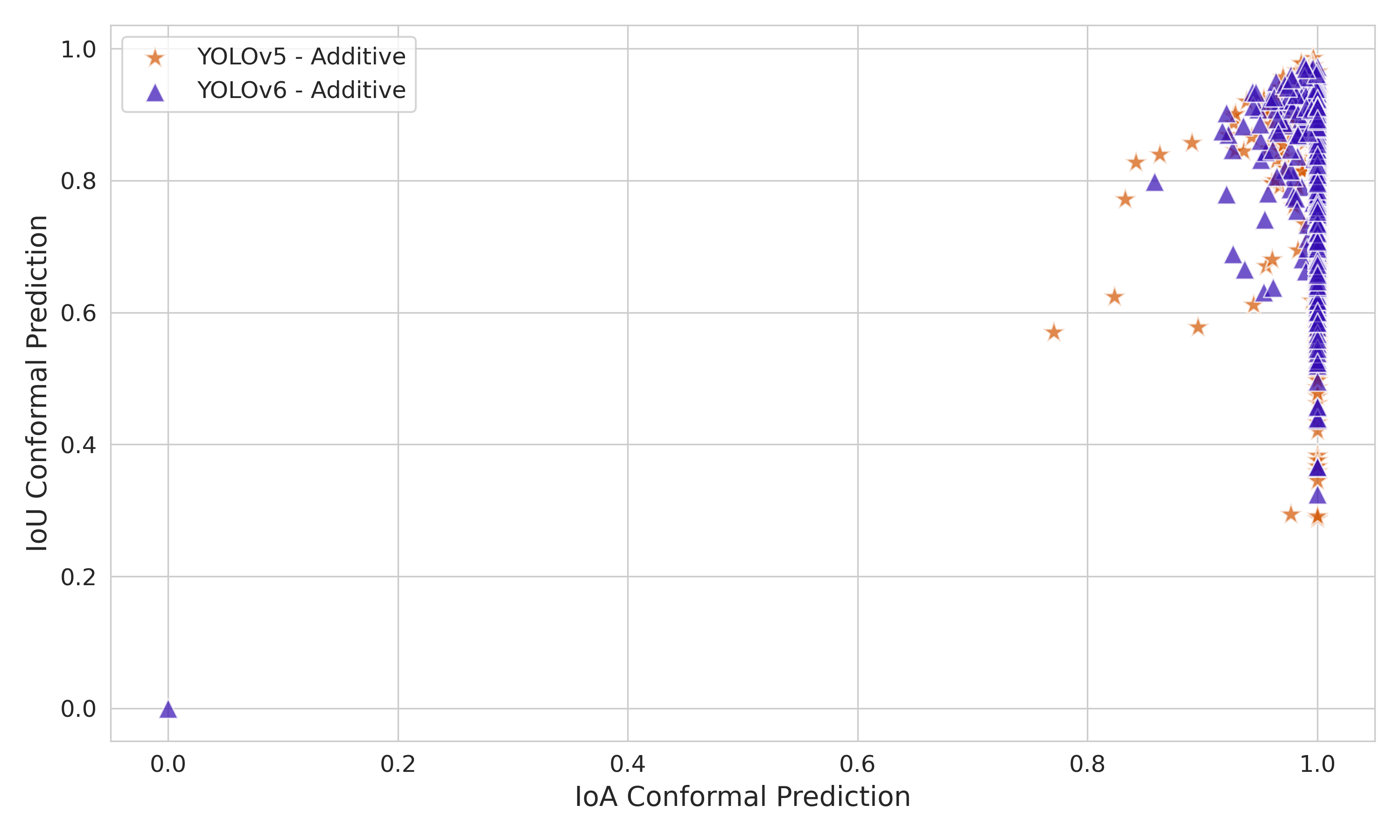}
  }
\end{figure}

Figure~\ref{fig:comparison_ioa_iou_conformal} shows the relationship between Intersection over Union (IoU) and Intersection over Area (IoA) for conformalized YOLO models. It highlights that the majority of conformalized predicted boxes, with more than 70\% (according to a user-defined ratio \(\alpha = 0.3\)), saturate at IoA = 1. This means that, for most predictions, the conformalized models meet the full coverage criterion (IoA = 1), which is a key feature of the conformalization process. Consequently, at most 30\% of the predicted boxes are likely to result in false negatives, as they fail to fully cover the ground truth. This behavior explains why the C-mAP scores for the conformalized models are consistently lower than 70\%, despite their improvement over standard YOLO models. To achieve higher C-mAP scores, users can decrease the value of \(\alpha\), which helps reduce the number of false positives as measured in C-mAP.

Figure~\ref{fig:comparison_iou_iou_conformal} illustrates a slight degradation of IoU in the conformalized versions of YOLO, both for the additive and multiplicative penalty variants. Some of the conformalized predicted boxes have IoU values that drop below the threshold of 0.5 after conformalization, leading to an increase in false positives. This degradation results in a reduction in both mAP and C-mAP scores. Overall, c-YOLOv5-a provides the best balance between high coverage, controlled bounding box and C-mAP.

\begin{figure}[h]
\floatconts
  {fig:comparison_iou_iou_conformal} 
  {\caption{Scatter plots comparing the Intersection over Union (IoU) of YOLO'S (x-axis) with their conformalized versions. Left: multiplicative penalty. Right: additive penalty.
}} 
  { 
    \includegraphics[width=0.48\linewidth]{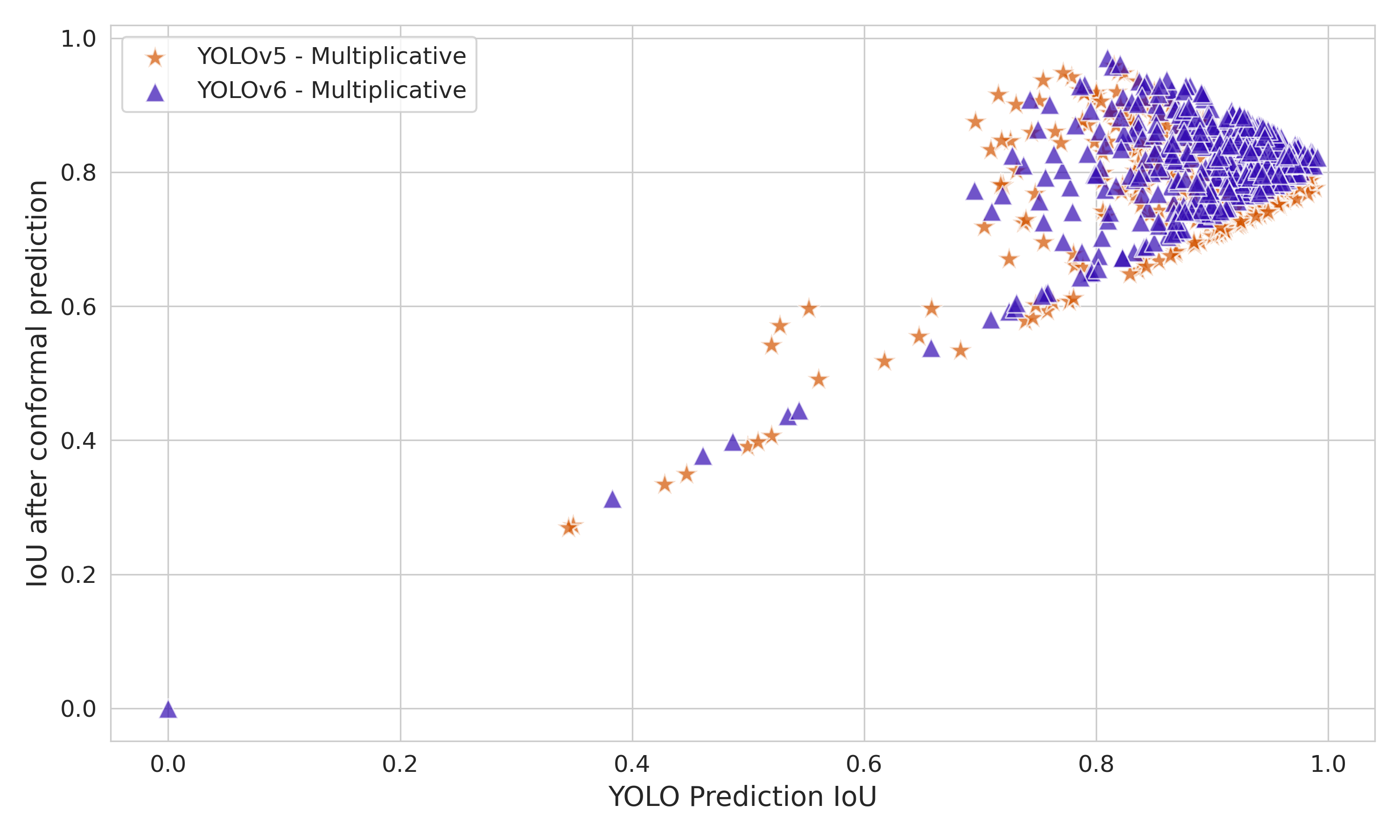}
    \includegraphics[width=0.48\linewidth]{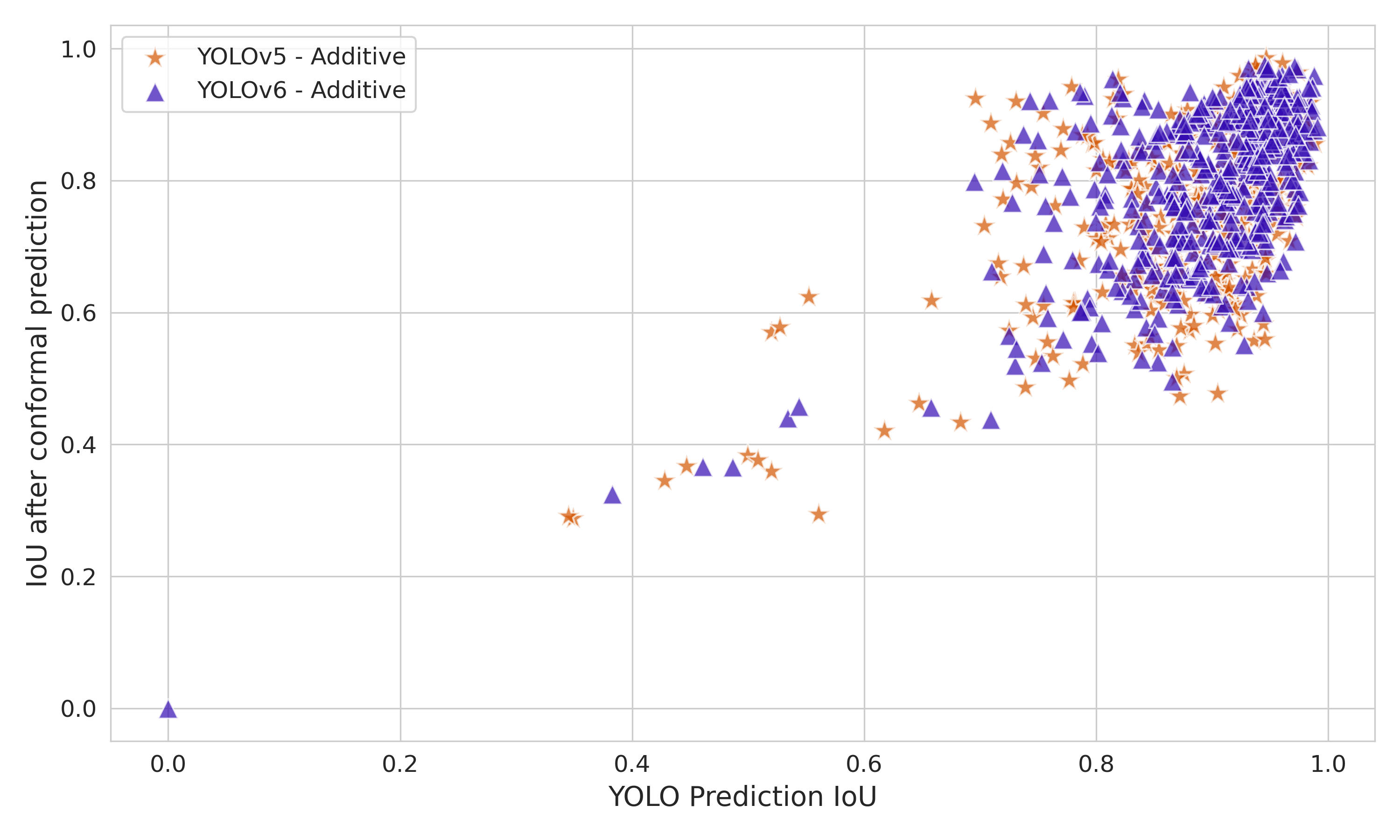}
  }
\end{figure}

\section{Conclusion}
\label{sec:conclusion}

Conformal Prediction (CP) offers a compelling alternative for uncertainty quantification. Unlike Bayesian or ensemble methods, CP requires no changes to the training process and provides formal, distribution-free guarantees. It can be applied to existing models, making it a practical and theoretically sound choice—especially in safety-critical applications where robustness and traceability are essential.

Conformal prediction delivers reliable uncertainty estimates with minimal overhead. However, it is not without limitations: it can produce overly conservative predictions—especially with corrections like Bonferroni adjustments—and its guarantees rely on calibration distributions matching test conditions. In multi-object detection, naive application can lead to artifacts like inflated bounding boxes or false positives. In practice, hybrid approaches—such as applying conformalization to the outputs of Bayesian models or deep ensembles—can combine the strengths of both paradigms, offering rich uncertainty estimates alongside formal, distribution-free coverage guarantees.

Finally, our current evaluation is limited by the LARD dataset, which contains only a single object per image. This restricts the analysis of the behavior of CP in multi-object scenes, where independent conformalization may lead to overlapping predictions and distort performance metrics such as mAP. Future work should investigate the impact of CP in more complex scenarios to fully assess its potential for dense detection tasks.

\acks{}
Our work has benefitted from the AI Interdisciplinary Institute ANITI. ANITI is funded by the France 2030 program under the Grant agreement n°ANR-23-IACL-0002.
The authors thank all the people and industrial partners involved in the DEEL project. This work has benefited from the support of the DEEL project,\footnote{\url{https://www.deel.ai/}} with fundings from the Agence Nationale de la Recherche, and which is part of the ANITI AI cluster.

\bibliography{pmlr-sample}

\begin{thebibliography}{26}
\providecommand{\natexlab}[1]{#1}
\providecommand{\url}[1]{\texttt{#1}}
\expandafter\ifx\csname urlstyle\endcsname\relax
  \providecommand{\doi}[1]{doi: #1}\else
  \providecommand{\doi}{doi: \begingroup \urlstyle{rm}\Url}\fi

\bibitem[And{\'e}ol et~al.(2023)And{\'e}ol, Fel, De~Grancey, and Mossina]{andeol2023confident}
L{\'e}o And{\'e}ol, Thomas Fel, Florence De~Grancey, and Luca Mossina.
\newblock Confident object detection via conformal prediction and conformal risk control: an application to railway signaling.
\newblock In \emph{Conformal and Probabilistic Prediction with Applications}, pages 36--55. PMLR, 2023.

\bibitem[Angelopoulos et~al.(2024)Angelopoulos, Bates, Fisch, Lei, and Schuster]{angelopoulos2022conformal}
Anastasios~N Angelopoulos, Stephen Bates, Adam Fisch, Lihua Lei, and Tal Schuster.
\newblock Conformal risk control.
\newblock In \emph{The Twelfth International Conference on Learning Representations}, 2024.

\bibitem[Balduzzi et~al.(2021)Balduzzi, Ferrari~Bravo, Chernova, Cruceru, van Dijk, de~Lange, Jerez, Koehler, Koerner, Perret-Gentil, et~al.]{daedalean}
Giovanni Balduzzi, Martino Ferrari~Bravo, Anna Chernova, Calin Cruceru, Luuk van Dijk, Peter de~Lange, Juan Jerez, Nathana{\"e}l Koehler, Mathias Koerner, Corentin Perret-Gentil, et~al.
\newblock Neural network based runway landing guidance for general aviation autoland.
\newblock Technical report, United States. Department of Transportation. Federal Aviation Administration~…, 2021.

\bibitem[Bharati and Pramanik(2020)]{bharati2020deep}
Puja Bharati and Ankita Pramanik.
\newblock Deep learning techniques—r-cnn to mask r-cnn: a survey.
\newblock \emph{Computational Intelligence in Pattern Recognition: Proceedings of CIPR 2019}, pages 657--668, 2020.

\bibitem[Carion et~al.(2020)Carion, Massa, Synnaeve, Usunier, Kirillov, and Zagoruyko]{carion2020end}
Nicolas Carion, Francisco Massa, Gabriel Synnaeve, Nicolas Usunier, Alexander Kirillov, and Sergey Zagoruyko.
\newblock End-to-end object detection with transformers.
\newblock In \emph{European conference on computer vision}, pages 213--229. Springer, 2020.

\bibitem[Copley et~al.(2024)Copley, Finlay, and Hiett]{copley2024uncertain}
Vicky Copley, Greg Finlay, and Ben Hiett.
\newblock The uncertain object: Application of conformal prediction to aerial and satellite images.
\newblock In \emph{The 13th Symposium on Conformal and Probabilistic Prediction with Applications}, pages 73--89. PMLR, 2024.

\bibitem[Dai et~al.(2024)Dai, Zhai, Wang, Zu, Shen, Lv, Lu, and Wang]{dai2024yomo}
Wei Dai, Zhengjun Zhai, Dezhong Wang, Zhaozi Zu, Siyuan Shen, Xinlei Lv, Sheng Lu, and Lei Wang.
\newblock Yomo-runwaynet: A lightweight fixed-wing aircraft runway detection algorithm combining yolo and mobilerunwaynet.
\newblock \emph{Drones}, 8\penalty0 (7):\penalty0 330, 2024.

\bibitem[de~Grancey et~al.(2022)de~Grancey, Adam, Alecu, Gerchinovitz, Mamalet, and Vigouroux]{de2022object}
Florence de~Grancey, Jean-Luc Adam, Lucian Alecu, S{\'e}bastien Gerchinovitz, Franck Mamalet, and David Vigouroux.
\newblock Object detection with probabilistic guarantees.
\newblock In \emph{Fifth International Workshop on Artificial Intelligence Safety Engineering (WAISE 2022)}, 2022.

\bibitem[Ducoffe et~al.(2020)Ducoffe, Gerchinovitz, and Gupta]{ducoffe2020high}
M{\'e}lanie Ducoffe, S{\'e}bastien Gerchinovitz, and Jayant~Sen Gupta.
\newblock A high probability safety guarantee for shifted neural network surrogates.
\newblock In \emph{SafeAI@ AAAI}, pages 74--82, 2020.

\bibitem[Ducoffe et~al.(2023)Ducoffe, Carrere, F{\'e}liers, Gauffriau, Mussot, Pagetti, and Sammour]{ducoffe2023lard}
M{\'e}lanie Ducoffe, Maxime Carrere, L{\'e}o F{\'e}liers, Adrien Gauffriau, Vincent Mussot, Claire Pagetti, and Thierry Sammour.
\newblock Lard--landing approach runway detection--dataset for vision based landing.
\newblock \emph{arXiv preprint arXiv:2304.09938}, 2023.

\bibitem[Ernez et~al.(2023)Ernez, Arnold, Galametz, Kobus, and Ould-Amer]{ernez2023applying}
Fares Ernez, Alexandre Arnold, Audrey Galametz, Catherine Kobus, and Nawal Ould-Amer.
\newblock Applying the conformal prediction paradigm for the uncertainty quantification of an end-to-end automatic speech recognition model (wav2vec 2.0).
\newblock In \emph{Conformal and Probabilistic Prediction with Applications}, pages 16--35. PMLR, 2023.

\bibitem[Harakeh et~al.(2020)Harakeh, Smart, and Waslander]{BayesOD}
Ali Harakeh, Michael Smart, and Steven~L Waslander.
\newblock Bayesod: A bayesian approach for uncertainty estimation in deep object detectors.
\newblock In \emph{2020 IEEE International Conference on Robotics and Automation (ICRA)}, pages 87--93. IEEE, 2020.

\bibitem[Jocher et~al.(2020)Jocher, Stoken, Borovec, Changyu, Hogan, Diaconu, Poznanski, Yu, Rai, Ferriday, et~al.]{jocher2020ultralytics}
Glenn Jocher, Alex Stoken, Jirka Borovec, Liu Changyu, Adam Hogan, Laurentiu Diaconu, Jake Poznanski, Lijun Yu, Prashant Rai, Russ Ferriday, et~al.
\newblock ultralytics/yolov5: v3. 0.
\newblock \emph{Zenodo}, 2020.

\bibitem[Li et~al.(2022)Li, Li, Jiang, Weng, Geng, Li, Ke, Li, Cheng, Nie, et~al.]{li2022yolov6}
Chuyi Li, Lulu Li, Hongliang Jiang, Kaiheng Weng, Yifei Geng, Liang Li, Zaidan Ke, Qingyuan Li, Meng Cheng, Weiqiang Nie, et~al.
\newblock Yolov6: A single-stage object detection framework for industrial applications.
\newblock \emph{arXiv preprint arXiv:2209.02976}, 2022.

\bibitem[Lyu et~al.(2020)Lyu, Gutierrez, Rajguru, and Beksi]{DeepEnsembles}
Zongyao Lyu, Nolan Gutierrez, Aditya Rajguru, and William~J Beksi.
\newblock Probabilistic object detection via deep ensembles.
\newblock In \emph{European Conference on Computer Vision}, pages 67--75. Springer, 2020.

\bibitem[Massena et~al.(2025)Massena, And{\'e}ol, Boissin, Friedrich, Mamalet, Serrurier, and Gerchinovitz]{massena2025efficient}
Thomas Massena, L{\'e}o And{\'e}ol, Thibaut Boissin, Corentin Friedrich, Franck Mamalet, Mathieu Serrurier, and S{\'e}bastien Gerchinovitz.
\newblock Efficient robust conformal prediction via lipschitz-bounded networks.
\newblock 2025.

\bibitem[Mendil et~al.(2023)Mendil, Mossina, and Vigouroux]{mendil2023puncc}
Mouhcine Mendil, Luca Mossina, and David Vigouroux.
\newblock Puncc: a python library for predictive uncertainty calibration and conformalization.
\newblock In \emph{Conformal and Probabilistic Prediction with Applications}, pages 582--601. PMLR, 2023.

\bibitem[Miller et~al.(2018)Miller, Nicholson, Dayoub, and S{\"u}nderhauf]{DropoutSampling}
Dimity Miller, Lachlan Nicholson, Feras Dayoub, and Niko S{\"u}nderhauf.
\newblock Dropout sampling for robust object detection in open-set conditions.
\newblock In \emph{2018 IEEE International Conference on Robotics and Automation (ICRA)}, pages 3243--3249. IEEE, 2018.

\bibitem[Neubeck and Van~Gool(2006)]{neubeck2006efficient}
Alexander Neubeck and Luc Van~Gool.
\newblock Efficient non-maximum suppression.
\newblock In \emph{18th international conference on pattern recognition (ICPR'06)}, volume~3, pages 850--855. IEEE, 2006.

\bibitem[Szegedy et~al.(2014)Szegedy, Zaremba, Sutskever, Bruna, Erhan, Goodfellow, and Fergus]{szegedy2013intriguing}
Christian Szegedy, Wojciech Zaremba, Ilya Sutskever, Joan Bruna, Dumitru Erhan, Ian Goodfellow, and Rob Fergus.
\newblock Intriguing properties of neural networks.
\newblock In \emph{The Second International Conference on Learning Representations}, 2014.

\bibitem[Terven et~al.(2023)Terven, C{\'o}rdova-Esparza, and Romero-Gonz{\'a}lez]{terven2023comprehensive}
Juan Terven, Diana-Margarita C{\'o}rdova-Esparza, and Julio-Alejandro Romero-Gonz{\'a}lez.
\newblock A comprehensive review of yolo architectures in computer vision: From yolov1 to yolov8 and yolo-nas.
\newblock \emph{Machine learning and knowledge extraction}, 5\penalty0 (4):\penalty0 1680--1716, 2023.

\bibitem[Tian et~al.(2020)Tian, Shen, Chen, and He]{tian2020fcos}
Zhi Tian, Chunhua Shen, Hao Chen, and Tong He.
\newblock Fcos: A simple and strong anchor-free object detector.
\newblock \emph{IEEE transactions on pattern analysis and machine intelligence}, 44\penalty0 (4):\penalty0 1922--1933, 2020.

\bibitem[Timans et~al.(2024)Timans, Straehle, Sakmann, and Nalisnick]{timans2024adaptive}
Alexander Timans, Christoph-Nikolas Straehle, Kaspar Sakmann, and Eric Nalisnick.
\newblock Adaptive bounding box uncertainties via two-step conformal prediction.
\newblock In \emph{European Conference on Computer Vision}, pages 363--398. Springer, 2024.

\bibitem[Vovk et~al.(2005)Vovk, Gammerman, and Shafer]{vovk2022algorithmic}
Vladimir Vovk, Alexander Gammerman, and Glenn Shafer.
\newblock \emph{Algorithmic learning in a random world}, volume~29.
\newblock Springer, 2005.

\bibitem[Wenkel et~al.(2021)Wenkel, Alhazmi, Liiv, Alrshoud, and Simon]{wenkel2021confidence}
Simon Wenkel, Khaled Alhazmi, Tanel Liiv, Saud Alrshoud, and Martin Simon.
\newblock Confidence score: The forgotten dimension of object detection performance evaluation.
\newblock \emph{Sensors}, 21\penalty0 (13):\penalty0 4350, 2021.

\bibitem[Yelleni et~al.(2024)Yelleni, Kumari, et~al.]{MCDropBlock}
Sai~Harsha Yelleni, Deepshikha Kumari, et~al.
\newblock Monte carlo dropblock for modeling uncertainty in object detection.
\newblock \emph{Pattern Recognition}, 146:\penalty0 110003, 2024.

\end{thebibliography}



\end{document}